\documentclass[pdflatex,sn-mathphys-num]{sn-jnl}% Math and Physical Sciences 
\usepackage{booktabs}
\usepackage{graphicx}
\usepackage{float}
\usepackage{multirow}%
\usepackage{amsmath,amssymb,amsfonts}%
\usepackage{amsthm}%
\usepackage{array}
\usepackage{mathrsfs}%
\usepackage[title]{appendix}%
\usepackage{xcolor}%
\usepackage{textcomp}%
\usepackage{manyfoot}%
\usepackage{algorithm}%
\usepackage{algpseudocode}%
\algtext*{EndIf}
\algtext*{EndFor}
\algtext*{EndWhile}
\usepackage{listings}%
\usepackage{lineno}
\usepackage{tikz}
\usepackage{pgfplots}
\usepackage{subcaption}
\usepackage{threeparttable}
\theoremstyle{thmstyleone}%
\theoremstyle{thmstyletwo}%
\theoremstyle{thmstylethree}%
\usepackage{siunitx}
\usepackage{caption}
\begin{document}
\title[ActOR]{ActOR: Ambient AI-Based Agentic Orchestration for Humanoid Autonomy in Operating Rooms}
\author*[1]{Han Zhang} \email{hzhan206@jhu.edu}\equalcont{Equal Contribution}
\author[1]{Sicheng Lu} \equalcont{Equal Contribution}
\author[1]{Aleks Santari} 
\author[2]{Jeffrey K. Jopling} 
\author[2]{Masaru Ishii} 
\author[1]{Mathias Unberath}
\affil[1]{\orgname{Johns Hopkins University},
\city{Baltimore}, \state{MD}, \country{USA}}
\affil[2]{\orgname{Johns Hopkins Medical Institutions},
\city{Baltimore}, \state{MD}, \country{USA}}

\abstract{
\textbf{Purpose:} Operating rooms (ORs) increasingly contain ambient sensing infrastructure in hopes that AI-based scene understanding would enable teams to optimize workflows and utilization. We pursue an alternative path: using ambient AI to orchestrate and inform robot action to benefit usage optimization. Deploying robots, such as humanoids, in the OR without ambient AI is severely limited by cumbersome de novo orientation and planning due to partial onboard observations. We posit that this can be achieved by coupling an ambient AI-based agentic orchestrator with the humanoid robotic actor, providing shared perception and execution feedback for humanoid autonomy in the OR.

\textbf{Methods:} ActOR fuses a wall-mounted ambient camera and the robot's head-mounted camera into a shared object-level digital twin (DT). A large language model agent plans over the DT using pre-programmed locomotion and manipulation skills without task-specific training. During execution, DT updates and execution outcomes inform retries, alternative approaches, or replanning.

\textbf{Results:} We evaluated ActOR in simulated and physical environments. Across three tasks in two simulated ORs reconstructed from real rooms, ambient perception reduced exploration from 1.05 to 0.12 \texttt{Explore} calls per trial when targets were initially outside the robot's view. Closed-loop feedback increased success from 46.7\% to 83.3\% under nominal execution and from 8.3\% to 51.7\% under injected navigation, object-relocation, and grasp perturbations. In a physical mock OR with a humanoid, ActOR completed 14 of 20 single-object disposal trials and demonstrated replanning after scene changes.

\textbf{Conclusion:} Ambient AI-based observations provide scene state that improves robot action planning, increasing success and reducing exploration compared with robot-only observations. Feeding skill outcomes and DT updates back to the planner improves multi-step task success. Our results show the feasibility of humanoid task autonomy for multi-step disposal tasks in a mock OR.
}

\maketitle
\section{Introduction}

Operating rooms (ORs) face a growing shortage of perioperative nursing and support staff~\cite{Ada_shortage}. Room turnover, supply retrieval, and disposal of used materials consume substantial staff time and directly affect OR efficiency~\cite{10.1159/000541659}. These tasks are repetitive and physically demanding, and do not require highly trained personnel. Automating these tasks with robots rather than relying on an already strained workforce is a promising avenue to improve turnover times. The postoperative phase provides a practical setting for automation: the patient has left, most staff have stepped away, and the environment is comparatively controlled~\cite{kernbach_humanoid_2026, wachs2026finalreportworkshoprobotics}. Humanoid robots are attractive for such tasks because they can, in principle, operate within spaces designed for humans without major modifications.

At the same time, modern ORs are becoming increasingly instrumented. Ceiling- and wall-mounted cameras have been used to recognize surgical phases, localize staff and equipment, and construct digital twins (DTs) of the OR~\cite{zhang_egosurg_2026,perez_privacy-preserving_2025,ozsoy_specialized_2026,wang_topoor_2026}. Yet this ambient sensing infrastructure has primarily been used to inform clinical teams rather than directly support robotic autonomy. We ask whether the instrumented OR itself can instead become part of the robot's perceptual infrastructure, providing persistent room-scale context that complements the robot's local, egocentric view.

This question is particularly relevant for humanoid autonomy in the OR, where two limitations make robot-centric operation difficult. First, autonomous task planning requires an accurate representation of the current environment, but an onboard camera provides only a moving and frequently occluded view of a cluttered room. Upon entering the room to complete an assigned task, a robot may therefore need to explore the environment before it can decide where and how to act. Second, humanoid execution remains imperfect: navigation can terminate short of a target, and grasps can fail. A plan generated from the initial scene state therefore becomes unreliable when execution deviates from expectation. Foundation-model agents can decompose open-ended instructions into robot skills~\cite{huh2026homebody}, but reliable physical autonomy also requires broader access to the world state and feedback about executed actions.

\begin{figure}[t]
    \centering
    \includegraphics[width=\textwidth]{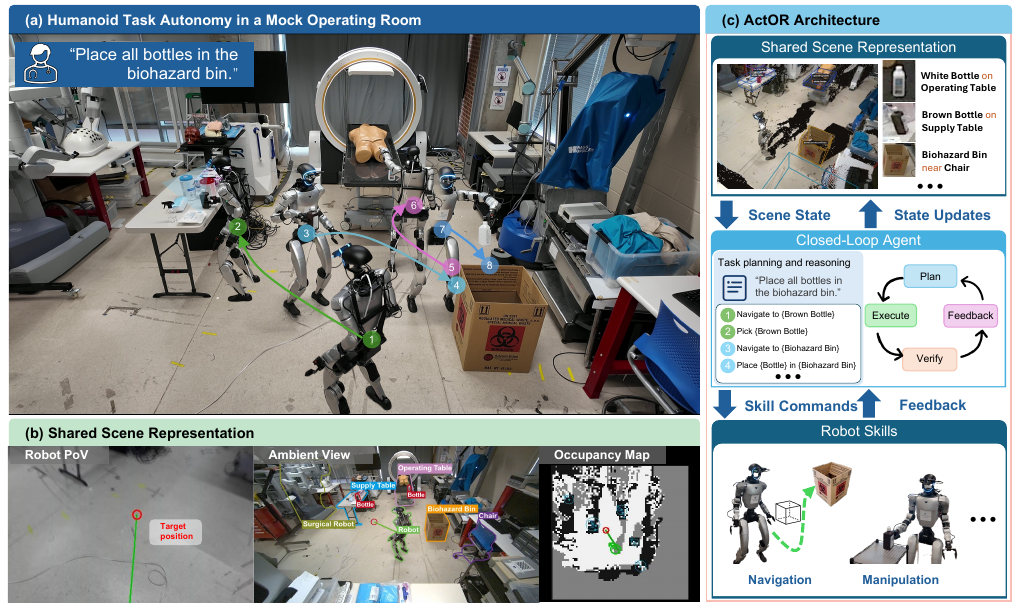}
    \caption{Overview of ActOR. (a) A humanoid executes a multi-object disposal task in a mock OR. (b) The shared scene representation integrates ambient and robot observations into an object-level DT and occupancy map. (c) An LLM agent uses the DT and robot skills for task planning, execution monitoring, and replanning.}    
    \label{fig:teaser}
\end{figure}

We address these limitations with \textbf{ActOR}, an ambient AI-based agentic orchestration framework for humanoid autonomy in the OR (Fig.~\ref{fig:teaser}). ActOR fuses observations from a wall-mounted ambient camera and the robot's head-mounted camera into a shared object-level DT, combining room-scale awareness with near-field observations for manipulation. A large language model (LLM) agent plans over the DT using a library of pre-programmed skills, without task-specific training. During execution, the DT is updated and the outcome is fed back to the agent, which retries, changes its approach, or replans. We evaluate ActOR on postoperative support tasks in two simulated ORs reconstructed from real rooms and on a Unitree G1 humanoid in a physical mock OR. Our contributions are:

\begin{itemize}
    \item A shared scene representation integrating room-scale and egocentric sensing for robot orientation and task planning in ORs.
    \item A closed-loop execution mechanism that feeds skill outcomes and scene updates back to an LLM planner for recovery and replanning.
    \item Simulation ablations of ambient perception and execution feedback, together with physical validation in a mock OR.
\end{itemize}

\section{Related Work}

Hospital robots have so far focused on logistics outside the OR, such as mobile manipulators that deliver supplies to nursing units~\cite{collaborative_robots_for_nurses}, while humanoids in the OR have been studied mainly under teleoperation~\cite{cho_humanoid_2026,liang_vivo_2026,atar2025humanoidshospitalstechnicalstudy}. Outside medicine, fixed external cameras have been used to localize mobile robots and build shared semantic maps~\cite{bultmann2023external}. ActOR uses ambient AI-based scene understanding to support humanoid task planning in the OR and incorporates execution feedback for retry and replanning.

LLMs can decompose open-ended instructions into skill sequences when grounded in a scene representation, such as open-vocabulary or hierarchical 3D scene graphs~\cite{gu2024conceptgraphs,rana2023sayplan,byrd2026constrained}. Closed-loop variants feed back success detection or scene descriptions to recover from failures~\cite{huang2022inner}. These systems typically build their representation from onboard sensing and rely on exploration to discover the scene~\cite{huh2026homebody}. In the OR, however, exploration is undesirable: the room is cluttered with sensitive equipment, and unnecessary robot motion should be minimized. ActOR therefore obtains the scene state from the ambient camera up front, and we evaluate each component through targeted ablations.

\section{Method}
\subsection{Problem Formulation}

We consider a humanoid robot performing postoperative support tasks in an OR. Given a natural-language instruction $I$ (e.g., ``put all used supplies into the biohazard bin'') and a skill library $\mathcal{K}$, the robot must complete the task autonomously. The system maintains an object-level DT $\mathcal{S}_t = \{o_i\}_{i=1}^{N_t}$, where each node
\begin{equation}
    o_i = (l_i,\ \mathbf{p}_i,\ \mathbf{B}_i,\ h_i),
\end{equation} consists of a semantic label $l_i$, a 3D position $\mathbf{p}_i \in \mathbb{R}^3$ in a shared world frame, a 3D extent $\mathbf{B}_i$, and an interaction state $h_i \in \{\texttt{free}, \texttt{in-hand}\}$. As shown in Fig.~\ref{fig:teaser}c, ActOR couples this DT (Sec.~\ref{sec:dt}) with an LLM agent that plans and verifies over it (Sec.~\ref{sec:agent}) and a humanoid that executes the resulting skills (Sec.~\ref{sec:skills}).

\subsection{Shared Scene Representation}
\label{sec:dt}

The ambient camera provides persistent room-scale coverage independent of robot pose, while the robot camera provides accurate near-field geometry for manipulation. Given an RGB image from either camera, a vision-language model (VLM) \cite{openai_6luna} detects task-relevant objects with labels $l_i$; each 2D box prompts SAM2~\cite{ravi_sam_2025} for an instance mask, whose pixels are back-projected with stereo depth and transformed into the world frame to obtain $\mathbf{p}_i$ and $\mathbf{B}_i$. 

The DT is kept consistent with the scene as the robot acts. It is updated after each manipulation skill and on each \texttt{Observe} call, and only nodes inside the observing camera's view are modified. A detection is merged into the existing node it overlaps most when the two occupy the same volume; otherwise, a new node is created. A node is removed when consecutive depth observations find its volume unoccupied. Manipulation also changes object state: when the gripper stops before full closure, the grasped node is marked \texttt{in-hand} and follows the gripper until \texttt{Place}. Spatial relations such as \texttt{in}, \texttt{near} and \texttt{on} are derived from the object positions and extents and passed to the planner with the DT.

\subsection{Closed-Loop Agentic Orchestration}
\label{sec:agent}

A large language model (GPT-6.1 Sol~\cite{openai_6sol}) serves as the planner. Its system prompt specifies its role, OR domain knowledge, the skill library (Table~\ref{tab:skills}), and rules for handling failures. At each decision step, it receives the instruction $I$, the DT serialized as JSON, and the execution history. It then calls a skill, emits a multi-step plan, or declares the task complete or failed. Skills in a multi-step plan execute sequentially. On skill failure or plan completion, the planner receives the updated DT and execution history to determine the next action (Algorithm~\ref{alg:loop}).

Recovery is left to the planner, which is instructed to retry a failed skill and otherwise change its approach or skip the object. An episode ends when the planner declares the task complete or failed, or when the elapsed time reaches $T_\mathrm{timeout}$.

\begin{table}
\centering
\caption{Skill library and the condition under which each skill reports success.}
\label{tab:skills}
\scriptsize
\begin{tabular}{@{}p{3.0cm}p{5.5cm}p{4.5cm}@{}}
\toprule
Skill & Function & Success condition \\
\midrule
\texttt{Observe(target)} & Capture \texttt{target} and update the DT & -- \\
\texttt{Explore()} & Navigate to likely target locations & -- \\
\texttt{Navigate(target)} & Walk to a standoff pose facing \texttt{target} & $\mathrm{dist}_{xy}(\mathbf{p}_{\text{base}}, \mathbf{B}_{\texttt{target}}) \le d_{\text{reach}}$ \\
\texttt{Pick(target)} & Grasp \texttt{target} & $h_{\texttt{target}} = \texttt{in-hand}$ \\
\texttt{Place(target,dest)} & Release \texttt{target} at \texttt{dest} & $\mathbf{p}_{\texttt{target}}^{xy} \in
\mathbf{B}_{\texttt{dest}}^{xy}$ at release \\
\bottomrule
\end{tabular}
\end{table}

\begin{algorithm}
\caption{ActOR closed-loop execution}
\label{alg:loop}
\begin{algorithmic}[1]
\Require Instruction $I$, skill library $\mathcal{K}$, DT $\mathcal{S}$, timeout $T_{\text{timeout}}$
\State $\mathcal{H} \gets \emptyset$ \Comment{execution history}
\While{elapsed time $< T_{\text{timeout}}$}
    \State $d \gets \textsc{Planner}(I, \mathcal{S}, \mathcal{H}, \mathcal{K})$
    \If{$d$ declares the task complete or failed}
        \State \Return $d$
    \EndIf
    \For{each skill $a \in \textsc{Skills}(d)$}
        \State $\textit{success} \gets \textsc{Execute}(a)$
        \State $\mathcal{S} \gets \textsc{UpdateDT}(\mathcal{S})$;\; $\mathcal{H} \gets \mathcal{H} \cup \{(a, \textit{success})\}$
        \If{\textbf{not} \textit{success}} \textbf{break} \EndIf
    \EndFor
\EndWhile
\State \Return failure
\end{algorithmic}
\end{algorithm}

\subsection{Skill Implementation}
\label{sec:skills}

Upper-body manipulation and lower-body locomotion are decoupled. For \texttt{Navigate}, ambient depth is projected onto the floor plane to form a 2D occupancy map. A standoff pose facing the target within arm reach is computed, and A* plans waypoints that are converted to base velocity commands and tracked by an RL-based whole-body controller \cite{li2025amo}. For \texttt{Pick}, the arm closer to the target is selected, and AnyGrasp \cite{fang2023anygrasp} predicts grasp candidates from the robot camera point cloud, guided by the target’s bounding box. We retain AnyGrasp candidates and combine them with candidates generated by a 3D bounding-box heuristic adapted from \cite{10909193}. A motion planner built on Pinocchio \cite{carpentier2019pinocchio} checks candidates for kinematic feasibility and collision-free trajectories; feasible candidates are ranked by hand pose and motion magnitude to select the grasp to execute. The selected trajectory is converted to joint commands via inverse kinematics. \texttt{Place} follows the same pipeline, with a release pose above the destination’s position.

\section{Experiments}

\begin{figure}[t]
    \centering
    \includegraphics[width=\textwidth]{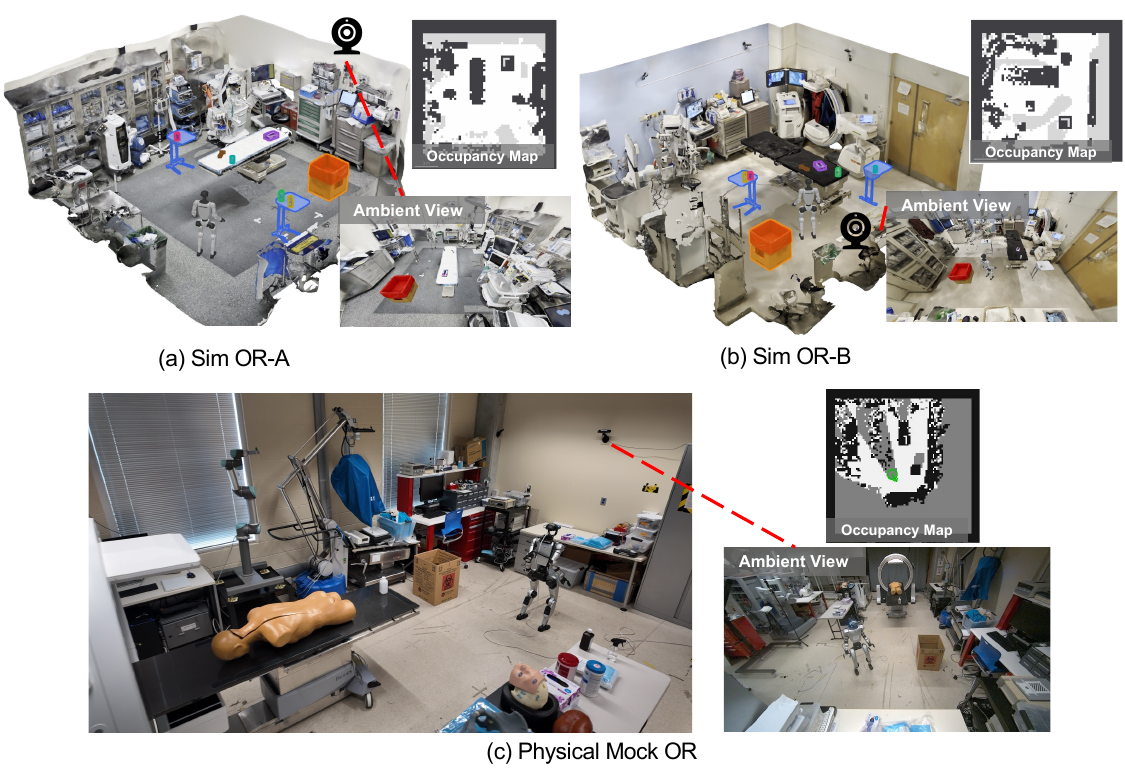}
    \caption{Experimental environments. (a, b) Simulated ORs reconstructed from real rooms for controlled studies. (c) Physical mock OR with a Unitree G1 humanoid.} 
    \label{fig:setup}
\end{figure}

We assume no human presence during execution, requiring the planner to resolve ambiguities without operator input. All task-relevant objects are placed within the robot’s reachable workspace. Controlled studies in MuJoCo~\cite{todorov2012mujoco} ablate ambient perception and closed-loop feedback, followed by physical deployment in a mock OR.

\subsection{Experimental Setup}

\paragraph{Environments}

Controlled studies were conducted in two simulated ORs with distinct layouts reconstructed from real environments~\cite{zhang_twinor_2026}, with object assets generated by SAM3D~\cite{sam3dteam2025sam3d3dfyimages}. Each environment included a virtual ambient camera. The simulated robot ran the same detection, segmentation, grasping, and skill stack as the physical system and executed through the whole-body controller in simulation; depth was rendered with simulated stereo noise. Physical experiments were conducted in a mock OR equipped with an operating table, mobile CT scanner, surgical robot, and supply tables.

\paragraph{Tasks}

We evaluated three simulation tasks: single-object retrieval, requiring delivery of a specified object; multi-object collection, requiring successive transfers of two objects; and room reset, requiring identification and disposal of task-relevant objects among distractors (e.g., ``put all used supplies into the biohazard bin''). Object layouts and initial robot poses were randomized across trials and kept identical across methods within each study.

A trial succeeded when all target objects reached the instructed destination within $T_\mathrm{timeout}$. We report task success rate (SR) over 60 trials per method and condition (3 tasks $\times$ 2 environments $\times$ 10 trials) and, for the ambient perception study, the mean number of \texttt{Explore} calls per trial. For physical experiments, we report end-to-end success and the source of each failure.

\paragraph{Implementation details}
We used a Unitree G1 EDU (29 DoF) with Dex1-1 grippers and a head-mounted ZED Mini, alongside a wall-mounted ZED X connected to a Jetson Orin. The shared world frame is defined by a checkerboard observed simultaneously by the ambient and head-mounted cameras. Hand--eye calibration registers the head-mounted camera to the robot frame, and the robot is localized in its map frame using onboard LiDAR SLAM. Composing these transformations at calibration time yields a fixed transformation from the SLAM map frame to the world frame. Depth estimation uses Fast-FoundationStereo~\cite{wen2026fastfoundationstereo}. GPT-6 Luna~\cite{openai_6luna} and GPT-6.1 Sol~\cite{openai_6sol} are accessed through API calls for detection and planning; other modules ran locally on an AMD Ryzen 9 5900HX and RTX 3080. We set $d_{\text{reach}} = 0.85\,\mathrm{m}$ and $T_{\text{timeout}} = 1800\,\mathrm{s}$.

\paragraph{Baselines}
We compare ActOR against two ablated variants. All methods share the same planner, perception modules, and skill implementations. 

\begin{itemize}
    \item \textbf{Robot-only.} The DT is built and updated only from the robot camera; execution remains closed-loop, and navigation uses the map built from the head camera.
    \item \textbf{Open-loop.} Same DT, planner, and prompt, but once execution has begun, the episode ends at the first skill that reports failure, removing retry and replanning.
    
\end{itemize}

\subsection{Controlled Simulation Study}

\begin{figure}[t]
    \centering
    \includegraphics[width=\linewidth]{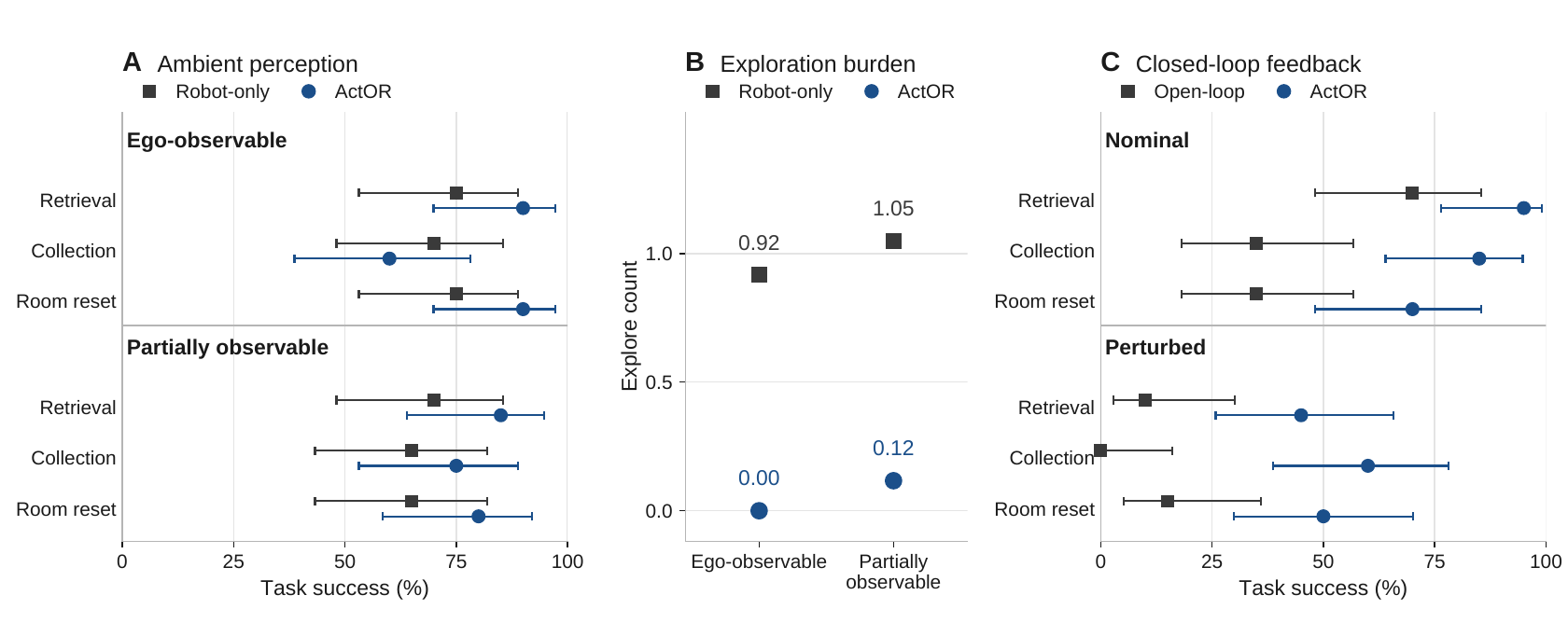}
    \caption{Controlled simulation results across two ORs. (a) Task success with and without ambient perception. (b) Mean \texttt{Explore} calls pooled across tasks. (c) Task success with and without closed-loop feedback. Error bars: Wilson 95\% confidence intervals.}
    \label{fig:sim_results}
\end{figure}

\paragraph{Ambient perception under partial observability}
We compared Robot-only and ActOR without injected perturbations (Fig.~\ref{fig:sim_results}a,b). For each object layout, the robot started either with all targets in the head camera's view (ego-observable) or with at least one target outside its view or occluded but visible to the ambient camera (partially observable).

When targets started outside the robot's view, Robot-only relied on exploration (1.05 \texttt{Explore} calls per trial), whereas ActOR localized targets from the ambient view and rarely explored (0.12 calls). Perception-related failures decreased from 13 to 1, and success was 80.0\% (48/60) for ActOR versus 66.7\% (40/60) for Robot-only. Under ego-observable conditions, success was 80.0\% (48/60) for ActOR and 73.3\% (44/60) for Robot-only. ActOR achieved higher success in both conditions, and the gap was larger when targets started outside the robot's view, consistent with the reduction in perception-related failures. Although the per-task confidence intervals overlap and a larger sample would be needed to confirm this gain, ambient perception nearly eliminated exploration in both conditions.

\paragraph{Closed-loop feedback under execution uncertainty}
We compared ActOR with Open-loop (Fig.~\ref{fig:sim_results}c) on a separate set of layouts whose initial robot poses were not controlled for head-camera observability; both methods used the same layouts and initial poses. We evaluated nominal execution and a perturbed condition in which each trial was assigned one of three perturbations: \textit{navigation error}, stopping 1.0--2.0\,m from the target; \textit{object relocation}, moving the target by at least 0.5\,m without updating the DT; or \textit{grasp misalignment}, offsetting the first grasp attempt by 5--8\,cm. Both methods were assigned the same perturbation on each trial.

Under nominal execution, ActOR succeeded in 83.3\% (50/60) of trials and Open-loop in 46.7\% (28/60). This gap indicates that individual skills fail often enough without injected perturbations that a single plan is unreliable, especially for multi-step tasks (Open-loop succeeded in 7/20 trials for both collection and room reset). Under perturbations, success dropped to 51.7\% (31/60) for ActOR and 8.3\% (5/60) for Open-loop; in multi-object collection, ActOR succeeded in 12/20 trials and Open-loop in none. Since Open-loop removes retry and replanning together, this gap reflects their combined effect.

By perturbation type, ActOR succeeded in 15/24 navigation-error trials, 11/18 object-relocation trials, and 5/18 grasp-misalignment trials, compared with 2/24, 3/18, and 0/18 for Open-loop (Supplementary Table S1). After a navigation error or an object relocation, the updated robot state and DT reflect the true configuration, so the planner can re-approach the target. A grasp misalignment is reported by \texttt{Pick} and retried without the offset, yet recovery was lowest for this type.

\subsection{Physical Mock OR Deployment}

\begin{figure}[t]
    \centering
    \includegraphics[width=\textwidth]{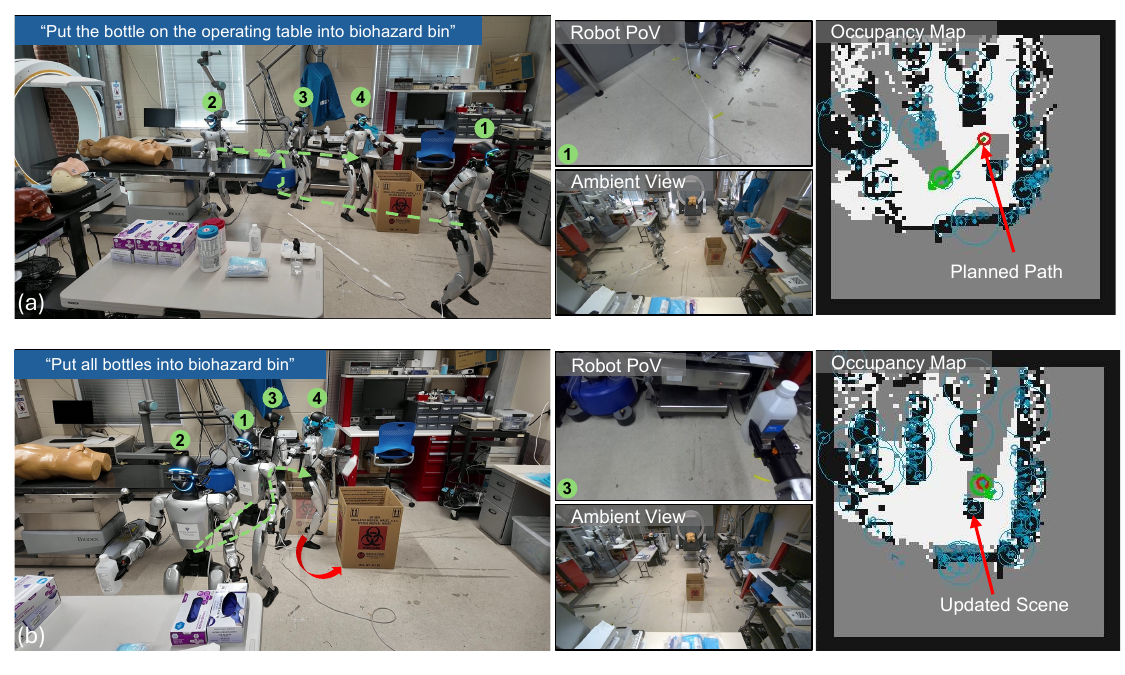}
    \caption{Representative humanoid autonomy in the mock operating room. (a) Multi-step execution in a single-object disposal trial. (b) Replanning after a scene change.}
    \label{fig:execution}
\end{figure}

We deployed the complete ActOR system on the Unitree G1 in the physical mock OR (Fig.~\ref{fig:setup}c). We conducted 20 single-object disposal trials, each instructing the robot to find a bottle and dispose of it in the biohazard bin. Bottle and bin placements varied across trials within the robot's reachable workspace. The robot was initialized with the bottle outside its egocentric view, so the bottle could initially be localized only through the ambient camera. A trial succeeded if the bottle reached the bin without human intervention. We also ran exploratory multi-object tests (``put all bottles into the biohazard bin''), including ones in which the scene was changed during execution by relocating the biohazard bin or introducing an additional bottle. 

ActOR succeeded in 14 of 20 single-object disposal trials (70\%; Fig.~\ref{fig:execution}a). In every successful trial, the bottle was localized from the ambient view and \texttt{Explore} was never invoked. Of the six failures, five originated in perception, where the VLM missed or misidentified the bottle, and one in execution, where a grasp still failed after the retry. In other trials, failed grasps were reported by the \texttt{Pick} skill and recovered by retrying. In the multi-object tests, ActOR completed multi-step executions and, in successful runs, replanned after scene changes toward the relocated bin or the added bottle (Fig.~\ref{fig:execution}b, Supplementary Video).

\section{Discussion}
Our results suggest that shared scene representation and closed-loop execution feedback together enable a humanoid to complete simple postoperative OR support tasks. The ambient camera provides room-level scene understanding, reducing the burden on the robot’s onboard sensing and exploration. The LLM agent composes pre-programmed skills without task-specific or room-specific training. More broadly, this work explores a path from ambient AI-based scene understanding to robot action in support of OR workflow and utilization optimization.

This study has several limitations. First, both detection and planning rely on proprietary models accessed through an API, which raises privacy concerns for clinical deployment. Locally deployable open-weight models offer a feasible alternative. For detection, however, even strong VLMs remain unreliable in cluttered scenes, and given the safety requirements of the OR, closed-vocabulary detectors trained on OR objects may be more appropriate. Second, the legged humanoid platform adds whole-body balance control on top of manipulation. This introduces additional constraints such as motor overheating and reduced manipulation accuracy, which currently limit the system to relatively large objects. A wheeled base would provide a more stable platform, at the cost of mobility in narrow spaces and over floor cables. Third, manipulation is limited to pick-and-place. More complex interactions, such as opening cabinets or pushing carts, require whole-body coordination. Learning-based policies combining imitation learning and reinforcement learning are a promising direction.

\section{Conclusion}
We presented ActOR, an ambient AI-based agentic orchestration framework integrating shared scene representation with closed-loop robot execution. Simulation studies demonstrated reduced exploration and improved task success under execution perturbations. Physical deployment demonstrated task execution and adaptation to scene changes, supporting the feasibility of humanoid autonomy for OR support tasks.
\backmatter

\bmhead{Code availability}
The code will be made publicly available upon publication.

\bmhead{Supplementary}
The video demonstrates humanoid autonomy in a mock OR.

\bmhead{Acknowledgments}
Supported by NSF CAREER Award No.~2239077.
\bibliography{references}

\clearpage
% \documentclass[pdflatex,sn-mathphys-num]{sn-jnl}% Math and Physical Sciences 
% \usepackage{booktabs}
% \usepackage{graphicx}
% \usepackage{multirow}%
% \usepackage{amsmath,amssymb,amsfonts}%
% \usepackage{amsthm}%
% \usepackage{array}
% \usepackage{mathrsfs}%
% \usepackage[title]{appendix}%
% \usepackage{xcolor}%
% \usepackage{textcomp}%
% \usepackage{manyfoot}%
% \usepackage{algorithm}%
% \usepackage{algorithmicx}%
% \usepackage{algpseudocode}%
% \usepackage{listings}%
% \usepackage{lineno}
% \usepackage{tikz}
% \usepackage{pgfplots}
% \usepackage{subcaption}
% \usepackage{threeparttable}
% \linenumbers
% \theoremstyle{thmstyleone}%
% \newtheorem{theorem}{Theorem}%  meant for continuous numbers
% \newtheorem{proposition}[theorem]{Proposition}% 
% \theoremstyle{thmstyletwo}%
% \newtheorem{example}{Example}%
% \newtheorem{remark}{Remark}%
% \theoremstyle{thmstylethree}%
% \newtheorem{definition}{Definition}%
% \usepackage{siunitx}
% \usepackage{mwe}
% \raggedbottom
% \newcommand{\tofill}[1]{\textcolor{red}{[#1]}}
% \newcommand{\tbd}{\textcolor{red}{xx}}
% \renewcommand{\thesection}{S\arabic{section}}
% \renewcommand{\thetable}{S\arabic{table}}
% \renewcommand{\thefigure}{S\arabic{figure}}
% \begin{document}

% \title[ActOR]{Supplementary Material for ActOR: Ambient AI-Based Agentic Orchestration for Humanoid Autonomy in Operating Rooms}

% \nolinenumbers
% \maketitle
% =====================================================================
\setcounter{section}{0}
\setcounter{table}{0}
\setcounter{figure}{0}
\setcounter{equation}{0}

\renewcommand{\thesection}{S\arabic{section}}
\renewcommand{\thetable}{S\arabic{table}}
\renewcommand{\thefigure}{S\arabic{figure}}
\renewcommand{\theequation}{S\arabic{equation}}

% Avoid duplicate hyperlink destinations
\renewcommand{\theHsection}{supp.\arabic{section}}
\renewcommand{\theHtable}{supp.\arabic{table}}
\renewcommand{\theHfigure}{supp.\arabic{figure}}
\renewcommand{\theHequation}{supp.\arabic{equation}}

\begin{center}
    {\Large\bfseries Supplementary Material}
\end{center}

\section{Simulation Results}
\label{sec:s_results}

\begin{table}[h]
\centering
\caption{Successful trials by perturbation type, pooled over tasks and environments. Denominators are the trials assigned each type.}
\label{tab:s_pert}
\small
\begin{tabular}{@{}lccc@{}}
\toprule
Method & Navigation error & Object relocation & Grasp misalignment \\
\midrule
Open-loop & 2/24 & 3/18 & 0/18 \\
ActOR     & 15/24 & 11/18 & 5/18 \\
\bottomrule
\end{tabular}
\end{table}

\begin{table}[h]
\centering
\caption{Successful trials by environment and task (out of 10).}
\label{tab:s_env}
\small
\begin{tabular}{@{}llcccccc@{}}
\toprule
& & \multicolumn{3}{c}{OR-A} & \multicolumn{3}{c}{OR-B} \\
\cmidrule(lr){3-5}\cmidrule(l){6-8}
Condition & Method & Retr. & Coll. & Reset & Retr. & Coll. & Reset \\
\midrule
\multirow{2}{*}{Ego-observable} & Robot-only & 7 & 7 & 6 & 8 & 7 & 9 \\
 & ActOR & 8 & 8 & 9 & 10 & 4 & 9 \\
\multirow{2}{*}{Partially observable} & Robot-only & 6 & 5 & 6 & 8 & 8 & 7 \\
 & ActOR & 8 & 8 & 8 & 9 & 7 & 8 \\
\midrule
\multirow{2}{*}{Nominal} & Open-loop & 7 & 3 & 2 & 7 & 4 & 5 \\
 & ActOR & 9 & 9 & 6 & 10 & 8 & 8 \\
\multirow{2}{*}{Perturbed} & Open-loop & 1 & 0 & 0 & 1 & 0 & 3 \\
 & ActOR & 4 & 6 & 4 & 5 & 6 & 6 \\
\bottomrule
\end{tabular}
\end{table}

\begin{table}[h]
\centering
\caption{Pooled success rates with 95\% Wilson confidence intervals.}
\label{tab:s_ci}
\small
\begin{tabular}{@{}llcc@{}}
\toprule
Condition & Method & Successes & SR [95\% CI] \\
\midrule
\multirow{2}{*}{Ego-observable} & Robot-only & 44/60 & 73.3 [61, 83] \\
 & ActOR & 48/60 & 80.0 [68, 88] \\
\multirow{2}{*}{Partially observable} & Robot-only & 40/60 & 66.7 [54, 77] \\
 & ActOR & 48/60 & 80.0 [68, 88] \\
\multirow{2}{*}{Nominal} & Open-loop & 28/60 & 46.7 [35, 59] \\
 & ActOR & 50/60 & 83.3 [72, 91] \\
\multirow{2}{*}{Perturbed} & Open-loop & 5/60 & 8.3 [4, 18] \\
 & ActOR & 31/60 & 51.7 [39, 64] \\
\bottomrule
\end{tabular}
\end{table}

% \paragraph{Perception-related failures.} A trial was labeled perception-related when, at the end of the episode, some target had no DT node within 0.30\,m of its true position, unless the robot fell or the episode timed out. The 13 and 1 such failures reported in the main text are for Robot-only and ActOR in the partially observable condition.

% \paragraph{Exploration under ego-observable starts.} Robot-only still invoked \texttt{Explore} in the ego-observable condition (0.92 calls per trial). Collection and room reset ask for every bottle in the room, and Robot-only's DT covers only the head camera's initial view, so the rest of the room remains unseen and the planner explores. ActOR's DT already includes the ambient view, leaving nothing to explore.
% =====================================================================
\section{Study Conditions}
\label{sec:s_conditions}

\paragraph{Tasks and objects.} Targets are drawn without replacement from four bottles (alcohol, saline, hydrogen peroxide, spray). Retrieval places one target; collection and room reset place two. Every layout also contains two distractors, a glove box and a reflex hammer, on a surface without targets. Room reset reuses the collection layout of the same seed and changes only the instruction (Table~\ref{tab:s_tasks}). The planner's site description identifies the bottles as the used disposable items (Sec.~\ref{sec:s_planner}).

\begin{table}[h]
\centering
\caption{Example instructions.}
\label{tab:s_tasks}
\small
\begin{tabular}{@{}lp{7cm}@{}}
\toprule
Task & Instruction \\
\midrule
Retrieval & ``Clean up: put the [appearance] into the biohazard box.'' \\
Collection & ``Put all the bottles into the biohazard box.'' \\
Room reset & ``Put all the used disposable items into the biohazard box.'' \\
\bottomrule
\end{tabular}
\end{table}

\paragraph{Observability conditions (ambient perception study).} Layouts place all targets on one surface, at most 0.5\,m apart; only the robot's start differs between conditions. In \textit{ego-observable} starts, the head camera sees every target fully and at least 30\% of the bin opening, at torso leans of 0$^\circ$ and 5$^\circ$; the robot starts within 1.2\,m of the targets' centroid. In \textit{partially observable} starts, at least one target is entirely outside the head camera's view or occluded. All targets remain visible to the ambient camera in both conditions. Visibility is tested by projecting sample points on each target into the head camera and ray-casting through the room's 2.5D height field.

\paragraph{Closed-loop study.} This study uses a separate set of layouts with one sampled start per layout, shared by both methods and both conditions. Starts are standable poses at least 0.8\,m from obstacles, with random heading, and are not controlled for head-camera observability.

\paragraph{Perturbations.} Perturbation types cycle with the trial index (4, 3, and 3 trials of the three types per task and room), and both methods are assigned the same type and random seed. Because the realized perturbation depends on execution (e.g., which object is approached first), it can differ between methods. Each perturbation fires at most once:
\begin{itemize}
  \item \textbf{Navigation error.} On the first navigation toward a graspable object, the robot stops 1.0--2.0\,m from it with up to 20$^\circ$ additional heading error, while \texttt{Navigate} reports success.
  \item \textbf{Object relocation.} At the start of that navigation, the target is moved at least 0.5\,m to another valid placement; the DT keeps its previous position until the object is re-observed.
  \item \textbf{Grasp misalignment.} On the first grasp, the grasp point is shifted 5--8\,cm along the finger direction. Retries are unperturbed.
\end{itemize}

% =====================================================================
\section{Implementation Details}
\label{sec:s_impl}
\paragraph{Simulation.} Both ORs were reconstructed from real rooms with TwinOR (OR-A: $5.46 \times 5.63$\,m; OR-B: $5.80 \times 5.42$\,m). The simulated G1 (29 DoF, Dex1-1 grippers) is controlled by the AMO whole-body policy in MuJoCo. The ambient camera renders $2560 \times 1440$ images (119$^\circ \times 88^\circ$ field of view, 120\,mm stereo baseline). The robot camera matches the physical head-mounted ZED Mini ($1280 \times 720$, about 85$^\circ \times 54^\circ$, 63\,mm baseline), pitched 46.6$^\circ$ down. Depth is MuJoCo-rendered depth with simulated stereo noise (0.15\,px disparity noise); detection, segmentation, and grasping use the same models as the physical system.

\paragraph{Shared scene representation.}
\begin{itemize}
  \item \textbf{Representation.} Object geometry is stored as 3\,cm voxels and serialized to the planner as a footprint radius and height range.
  \item \textbf{Association.} Two detections are merged when each voxel set covers at least 40\% of the other, or when one lies at least 70\% inside the other and their labels are compatible.
  \item \textbf{Updates.} The VLM runs on each \texttt{Observe} call (ambient and head cameras) and on the head camera before each \texttt{Pick} and \texttt{Place}. Depth-based occupancy and free-space carving run at about 1\,Hz.
  \item \textbf{Removal.} A voxel is deleted after two consecutive carving rounds observe free space through it; a node is removed when more than half of its tested voxels (at least ten) are empty, or fewer than five voxels remain. Held objects are not carved.
  \item \textbf{In-hand state.} After closing, the grasp is \texttt{in-hand} when the normalized gripper opening stops between 0.005 and 0.9; below 0.005 the fingers met and the grasp is empty.
  \item \textbf{Relations.} \texttt{on} and \texttt{in} are derived from vertical contact and footprint overlap with a supporting object; \texttt{near} links objects within 1.0\,m edge to edge (at most three neighbours).
\end{itemize}

\paragraph{Skills and planner.}
\begin{itemize}
  \item $d_\mathrm{reach} = 0.85$\,m is measured to the target's footprint boundary (0.60\,m planner clearance plus 0.25\,m). The occupancy grid has 0.10\,m resolution.
  \item Retry is a prompt rule, not a fixed loop. After a grasp failure, the planner calls \texttt{Pick} again from the same place; after an approach failure, it approaches from the side nearest an edge and tries once more. The same failure twice from the same place leads it to change the side, the object order, or the target, or to leave the object. Open-loop ends the episode at the first skill failure after execution has begun.
  \item $T_\mathrm{timeout} = 1800$\,s is a hard stop; the planner is warned at 1200\,s.
\end{itemize}

% =====================================================================
\section{Planner Prompt}
\label{sec:s_planner}

Excerpt of the system prompt used in OR-A, reproduced verbatim; [...] marks omitted passages. OR-B's SITE text differs in its room name and room description, in naming the procedure table instead of the operating table, and in its list of objects such a room typically holds. In the prompt, \texttt{scene\_graph} corresponds to \texttt{Observe}. The full prompt, including skill failure messages, will be released with the code.

\begin{lstlisting}[basicstyle=\ttfamily\scriptsize,breaklines=true,columns=fullflexible]
[...]
How to work:
- Look first. An empty or thin WORLD MODEL means nobody has looked yet, not an empty room: scene_graph with the smallest scope the instruction needs -- "robot", "targets" (one direct phrase per thing) or "full" for a whole-room task. SCENE GRAPH.cameras are the cameras that made that look; the head camera alone sees only what is ahead of the robot. explore, a walk past the room's landmarks looking with the head camera, only when those cameras leave unseen what the task needs. What WORLD MODEL already holds keeps its uid: to refresh its relations or reachability pass target_uids (no camera look); phrase in targets only what is not there yet, by its direct category or appearance, never the whole instruction. Two candidates you cannot tell apart: scene_graph scope "inspect". An inspect only settles a doubt that changes what you do next --not to confirm each object before acting on it: to tidy or move a thing you need where it is, not exactly what it is. A look that did not see it, from a camera that saw little of it unblocked, does not prove it is not there.
[...]
- A failed skill's note is your evidence; read it before the next move. The skills do not retry from elsewhere -- choosing what to try next is yours: * A pick that failed by the thing -- grasp: grasp_missed, not_lifted, no_reachable_grasp -- leaves the robot standing by it: pick again (the head camera measures it afresh and the robot finds its stance from there) -- not navigate, and no look first.
[...]
* dropped, dropped_on_the_way, target_has_no_position, not_seen_up_close, or the thing no longer where it was: look again (scene_graph, scope targets) before another try.
[...]
* The same failure twice from the same place will come a third time: change the side, the order or the thing, or leave it -- then finish the rest and report what was left and why.
- A multi-step task: one plan, then its steps in order, each call_skill carrying "plan_step": <id> at the top level. Moving things: a navigate step before every pick and every place. Give every step its full skill args: once you start step 1, each step that succeeds starts the next by itself, and you are asked again only when a step fails, the plan reaches a step no skill does, or it runs out -- then revise the plan from what is left, or declare. A step no skill can do (skill null: open, wipe ...) is re-emitted as skipped with what a person must do there -- never faked, never waited for.
[...]
- Declare succeeded as soon as the skill that completes the instruction reports succeeded; never call it again. A task with several things is succeeded only when every one is done; with any left, declare failed, the summary saying what was done and what is left, with why. A whole-room task is declared only after a scope "full" look made since your last pick or place: things the first look missed show up there.
[...]
- Calling a skill cancels the running one: "wait" while it works. "say" reports without stopping, sparingly. Nobody is there to answer a question: what a look cannot settle, decide yourself.

SITE: you work in a hospital operating room. The operating table stands in the middle under the surgical lights, the anaesthesia machine at its head, IV poles and equipment towers around it. Supply cabinets, carts, small stainless tables, waste bins and linen hampers line the walls. Wall-mounted things (sharps containers, dispensers, phone, clock, posters) are part of the walls, not things to fetch. From the room camera high on the wall a bottle on a table is only a few centimetres across: when a look misses one, look closer at the tops where small items are put down. Small items (bottles a hand tall, a glove box, a reflex hammer) are put down at the outer edge of the operating table's top and the Mayo stand trays, where a hand reaches them from the floor beside it. Only these are loose: anything else that looks like one -- a bottle, box or pack on a counter, shelf or cabinet top, on a cart with shelves or drawers, high on a wall, or inside a bin or hamper -- is fixed as part of the room, never what an instruction means and never to be moved. Where a thing stands is its measured place, not a relation's word: one at the edge of a Mayo tray or of the operating table's top is loose even when SCENE GRAPH has it in or next to a cabinet or cart beside that top (a big neighbour's box takes in what stands beside it). After a procedure the bottles put down there have been opened and used: they, and nothing else here, are the used disposable items, the used supplies and the waste an instruction may ask to clear away. The glove box (a full box of new gloves) and the reflex hammer (an instrument used again and again) are neither: they stay where they are. Nothing flat -- a mark, stain or patch on a top or a sheet -- is an item.
[...]
\end{lstlisting}

\end{document}